\documentclass[10pt]{article}
\usepackage{hyperref}
\usepackage{mathtools}
\usepackage{graphics}
\usepackage{float}
\usepackage{url}
\usepackage{wrapfig}
\usepackage{verbatim}
\usepackage{algpseudocode}
\usepackage[small]{caption}
\usepackage[round, authoryear]{natbib}
\usepackage{color,colortbl}
\usepackage{xcolor}
\usepackage{inputenc}
\usepackage{caption}
\usepackage{subcaption}
\usepackage{ctable}
\usepackage{amsmath}
\usepackage{amssymb}
\usepackage{bm}

\newcommand{\vb}{\mathbf}
\newcommand{\pmsep}{\hspace{1mm}$\pm$\hspace{1mm}}

\title{Discriminative training for Convolved Multiple-Output Gaussian processes}

\author{Sebasti\'an G\'omez-Gonz\'alez, Mauricio A. \'Alvarez, Hern\'an Felipe Garc\'ia\\
  {\small \emph{Faculty of Engineering, Universidad Tecnol\'ogica de Pereira,
      Colombia, 660003.}}\\}
  \date{}

\begin{document}
\maketitle

\begin{abstract}
Multi-output Gaussian processes (MOGP) are probability distributions
over vector-valued functions, and have been previously used for
multi-output regression and for multi-class classification. A less
explored facet of the multi-output Gaussian process is that it can be
used as a generative model for vector-valued random fields in the
context of pattern recognition.  As a generative model, the
multi-output GP is able to handle vector-valued functions with
continuous inputs, as opposed, for example, to hidden Markov
models. It also offers the ability to model multivariate random
functions with high dimensional inputs. In this report, we use a
discriminative training criteria known as Minimum Classification Error
to fit the parameters of a multi-output Gaussian process. We compare
the performance of generative training and discriminative training of
MOGP in emotion recognition, activity recognition, and face
recognition. We also compare the proposed methodology against hidden
Markov models trained in a generative and in a discriminative way.
\end{abstract}


\section{Introduction}
\label{sec:intro}

A growing interest within the Gaussian processes community in Machine
learning has been the formulation of suitable covariance functions for
describing multiple output processes as a joint Gaussian
process. Examples include the semiparametric latent factor model
\cite{Teh:semiparametric05}, the multi-task Gaussian process
\cite{Bonilla:multi07}, or the convolved multi-output Gaussian process
\cite{Boyle:dependent04,Alvarez:sparse2009}. Each of these methods
uses as a model for the covariance function either a version of the
linear model of coregionalization (LMC) \cite{Goovaerts:book97} or
a version of process convolutions (PC) \cite{Higdon:convolutions02}. Different
alternatives for building covariance functions for multiple-output
processes have been reviewed by
\cite{Alvarez:kernel:vector:valued:review:2012}.

Multiple output GPs have been used for supervised learning problems,
specifically, for multi-output regression \cite{Bonilla:multi07}, and
multi-class classification
\cite{Skolidis:multiclassICM2011,Chai:JMLR:2012}. The interest has
been mainly on exploiting the correlations between outputs to improve
the prediction performance, when compared to modeling each output
independently. In particular, a Gaussian process is used as a prior
over vector-valued functions $\mathbf{f}(\mathbf{x})$ mapping from
$\mathbf{x}\in\mathbb{R}^{p}$ to $\mathbf{f}\in
\mathbb{R}^{D}$. Components of $\mathbf{f}$ may be continuous or
discrete.

In this report, we advocate the use of multi-output GPs as generative
models for vector-valued random fields, this is, we use multi-output
GPs to directly modeling $p(\mathbf{f}(\mathbf{x}))$. Afterwards, we
use this probabilistic model to tackle a classification problem. An
important application area where this setup is of interest is in
multivariate time series classification. Here the vector-valued
function $\mathbf{f}$ is evaluated at discrete values of $\mathbf{x}$,
and it is typically modeled using an unsupervised learning method like
a hidden Markov model (HMM) or a linear dynamical system (LDS)
\cite{bishop2007}. Notice that by using a multi-output GP to model
$\mathbf{f}(\mathbf{x})$, we allow the vector-valued function
$\mathbf{f}(\mathbf{x})$ to be continuous on the input
space. Furthermore, we are able to model multivariate random functions
for which $p > 1$.  It is worth mentioning that the model we
propose here, is different from classical GP classification as
explained for example in \cite{Rasmussen:book06}. In standard GP
classification, the feature space is not assumed to follow a particular
structure, whereas in our model, the assumption is that the
feature space may be structured, with potentially correlated and
spatially varying features.


As a generative model, the multi-output Gaussian process can be used
for classification: we fit a multi-output GP for every class
independently, and to classify a new vector-valued random field, we
compute its likelihood for each class and make a decision using Bayes
rule. This generative approach works well when the real multivariate
signal's distribution is known, but this is rarely the case. Notice
that the optimization goal in the generative model is not a function
that measures classification performance, but a likelihood function
that is optimized for each class separately.

An alternative is to use discriminative training
\cite{Jebara:book:2004} for estimating the parameters of the
multi-output GP. A discriminative approach optimizes a function
classification performance directly. Thus, when the multi-output GP is
not an appropriate generative distribution the results of the
discriminative training procedure are usually better. There are
different criteria to perform discriminative training, including
maximum mutual information (MMI) \cite{Gopalakrishnan:MMI:1991}, and
minimum classification error (MCE) \cite{juangmce}.

In this report we present a discriminative approach to estimate the
hyperparameters of a multi-output Gaussian Process (MOGP) based on
minimum classification error (MCE). In section \ref{sec:methods} we
review how to fit the multi-output GP model using the generative
approach, and then we introduce our method to train the same MOGP
model with a discriminative approach based on MCE. In section
\ref{sec:results} we show experimental results, with both the
generative and discriminative approaches. Finally, we present
conclusions on section \ref{sec:conclusions}.

\section{Generative and discriminative training of multi-output GPs
}\label{sec:methods}

In our classification scenario, we have $M$ classes. We want to come up with
a classifier that allows us to map the matrix $\mathbf{F}(\mathbf{X})$ to one of the $M$
classes. Columns of matrix $\mathbf{X}$ are input vectors $\mathbf{x}_n\in\mathbb{R}^p$, and columns
of matrix $\mathbf{F}$ are feature vectors $\mathbf{f}({\mathbf{x}}_n) \in \mathbb{R}^D$, for some $n$ in an index set. Rows for $\mathbf{F}$ correspond
to different entries of $\mathbf{f}({\mathbf{x}}_n)$ evaluated for all $n$.
For example, in a multi-variate time series classification problem, $\mathbf{x}_n$ is a time point $t_n$, and $\mathbf{f}({\mathbf{x}}_n)$ is the multi-variate
time series at $\mathbf{x}_n=t_n$. Rows of the matrix $\mathbf{F}$ are the different time series.

The main idea that we introduce in this report is that we model the
class-conditional density $p(\mathbf{F}|\mathbf{X}, \mathcal{C}_m,
\bm{\theta}_m)$ using a multi-output Gaussian process, where
$\mathcal{C}_m$ is the class $m$, and $\bm{\theta}_m$ are
hyperparameters of the multi-output GP for class $m$. By doing so, we
allow correlations across the columns of $\mathbf{F}$, this is between
$\mathbf{f}({\mathbf{x}}_n)$ and $\mathbf{f}({\mathbf{x}}_m)$, for
$n\ne m$, and also allow correlations among the variables in the
vector $\mathbf{f}({\mathbf{x}}_n)$, for all $n$. We then estimate
$\bm{\theta}_m$ for all $m$ in a generative classification scheme, and
in a discriminative classification scheme using minimum classification
error. Notice that a HMM would model $p(\mathbf{F}| \mathcal{C}_m,
\bm{\theta}_m)$, since vectors $\mathbf{f}_n$ would be already defined
for discrete values of ${\mathbf{x}}$. Also notice that in standard GP
classification, we would model $p(\mathcal{C}_m|\mathbf{F})$, but with
now particular correlation assumptions over the entries in
$\mathbf{F}$.

Available data for each class are matrices $\mathbf{F}_m^l$, where
$m=1, \ldots, M$, and $l=1,\ldots, L_m$. Index $\l$ runs over the
instances for a class, and each class has $L_m$ instances. In turn, each matrix
$\mathbf{F}_m^l\in \mathbb{R}^{D\times N_m^l} $ with columns
$\mathbf{f}_m^l({\mathbf{x}_n})\in \mathbb{R}^D$, $\mathbf{x}_n\in
\mathbb{R}^p$, and $n=1, \ldots, N_m^l$.  To reduce clutter in the
notation, we assume that $L_m =L$ for all $m$, and $N_m^l = N$ for all
$m$, and $l$. Entries in $\mathbf{f}_m^l({\mathbf{x}_n})$ are given by
$f_{d}^{l, m}(\mathbf{x}_n)$ for $d=1, \ldots, D$. We define the
vector $\hat{\mathbf{f}}^{l,m}_d$ with elements given by $\{f_{d}^{l,
  m}(\mathbf{x}_n)\}_{n=1}^N$. Notice that the rows of $\mathbf{F}_m^l$
are given by $(\hat{\mathbf{f}}^{l,m}_d)^{\top}$. Also, vector
$\mathbf{f}^{l}_m=[(\hat{\mathbf{f}}^{l,m}_1)^{\top} \ldots (\hat{\mathbf{f}}^{l,m}_D)^{\top}]^{\top}$.
We use $\mathbf{F}_m$ to collectively refer to all matrices
$\{\mathbf{F}_m^l\}_{l=1}^{L}$, or all vectors $\{\mathbf{f}_m^l\}_{l=1}^{L}$.
We use $\mathbf{X}_{m}^l$ to refer to the set of input vectors
$\{\mathbf{x}_n\}_{n=1}^N$ for class $m$, and instance $l$. $\mathbf{X}_{m}$ refers to all the
matrices $\{\mathbf{X}_{m}^l\}_{l=1}^L$. Likewise, $\bm{\Theta}$ refers to the set $\{\bm{\theta}_m\}_{m=1}^M$.

\subsection{Multiple Outputs Gaussian Processes}\label{multigps}

According to \cite{Rasmussen:book06}, a Gaussian Process is a collection of random variables, any finite number of
which have a joint Gaussian distribution. We can use a Gaussian process to model a distribution over functions. Likewise, we can use a multi-output Gaussian process to model a distribution over vector-valued functions $\vb{f}(\vb{x}) = [f_1(\vb{x}) \ldots f_D(\vb{x})]^{\top}$.
The vector valued function $\vb{f}(\vb{x})$ is modeled with a GP,
\begin{align*}
\vb{f}(\vb{x})\sim\mathcal{GP}(\mathbf{0},\mathbf{k}(\mathbf{x}, \mathbf{x}')),
\end{align*}
where $\mathbf{k}(\mathbf{x}, \mathbf{x}')$ is a kernel for vector-valued functions \cite{Alvarez:kernel:vector:valued:review:2012}, with
entries given by $k_{f_d,f_{d'}}(\mathbf{x},\mathbf{x}')$. A range of alternatives for
$k_{f_d,f_{d'}}(\mathbf{x},\mathbf{x}')$ can be summarized using the general
expression
\begin{equation}
  \begin{split}
 \sum_{q=1}^Q\sum_{i=1}^{R_q} \int\int G^i_{d,q}(\mathbf{x},\mathbf{z})G^i_{d',q}(\mathbf{x'},\mathbf{z'})
  k_q(\mathbf{z},\mathbf{z}')d\mathbf{z'}d\mathbf{z},
  \end{split}
\label{eq:covf}
\end{equation}
where $Q$ is the number of latent functions used for constructing the kernel; $R_q$ is the number of latent functions (for a particular $q$)
sharing the same covariance; $G^i_{d,q}(\mathbf{x}-\mathbf{z})$ is known as the smoothing kernel for output $d$, and
$k_q(\mathbf{z},\mathbf{z}')$ is the kernel of each latent function $q$. For details, the reader is referred to
\cite{Alvarez:kernel:vector:valued:review:2012}. In the linear model of coregionalization,
$G^i_{d,q}(\mathbf{x},\mathbf{z}) = a^i_{d,q}\delta(\mathbf{x}-\mathbf{z})$,
where $a^i_{d,q}$ are constants, and $\delta(\cdot)$ is the Dirac delta function.

In all the experimental section, we use a kernel of the form \eqref{eq:covf} with $R_q=1$. We also assume that both
$G^i_{d,q}(\mathbf{x}, \mathbf{z})$ and $k_q(\mathbf{z},\mathbf{z}')$ are given by Gaussian kernels of the form
\begin{align*}
k(\mathbf{x}, \mathbf{x}') & = \frac{|\bm{\Lambda}|^{1/2}}{(2\pi)^{p/2}}
\exp\left[-\frac{1}{2}\left(\mathbf{x} - \mathbf{x}'\right)^{\top}\bm{\Lambda}\left(\mathbf{x} - \mathbf{x}'\right)\right],
\end{align*}
where $\bm{\Lambda}$ is the precision matrix.

Given a set of input vector $\mathbf{X} = \{\mathbf{x}_n\}_{n=1}^N$,
the columns of the matrix $\mathbf{F}$ correspond to the vector-valued
function $\vb{f}(\vb{x})$ evaluated at $\mathbf{X}$. Notice that the
rows in $\mathbf{F}$ are vectors $\hat{\mathbf{f}}_d^{\top} =
[f_d(\mathbf{x}_1) \ldots f_d(\mathbf{x}_N)]$. In multi-output GP, we model the vector
$\mathbf{f} = [\hat{\mathbf{f}}_1^{\top} \ldots \hat{\mathbf{f}}_D^{\top}]^{\top}$ by
$\vb{f} \sim \mathcal{N}(\vb{0},\vb{K})$, where $\vb{K}\in \mathbb{R}^{ND\times ND}$, and the entries in
$\vb{K}$ are computed using $k_{f_d,f_{d'}}(\mathbf{x}_n,\mathbf{x}_m)$, for all $d,d'=1,\ldots, D$, and
$n,m=1,\ldots, N$.

\subsection{Generative Training}
\label{subs:generative}

In the generative model, we train separately a multi-output GP for each class. In our
case, training consists of estimating the kernel hyperparameters of the
multi-output GP, $\bm{\theta}_m$. Let us assume that the training set
consists of several multi-output processes grouped in $\vb{F}_m$ and drawn
independently, from the Gaussian process generative model given by
\begin{equation*}
    p(\vb{f}_m^l|\mathbf{X}_m^l, \mathcal{C}_m, \bm{\theta}_m) =
\mathcal{N}(\vb{f}_m^l|\vb{0},\vb{K}_m),
\end{equation*}
where $\vb{K}_m$ is the kernel matrix for class $m$, as explained in section \ref{multigps}.

In order to train the generative model, we maximize the log marginal likelihood
function with respect to the parameter vector $\bm{\theta}_m$. As we assumed that the different
instances of the multi-output process are generated independently given the kernel
hyperparameters, we can write the log marginal likelihood for class $m$,
$\log(p(\vb{F}_m|\bm{\theta}_m))$, as
\begin{align}
 -\frac{1}{2}\sum_{l=1}^L\left[\vb{f}_l^\top\vb{K}_m^{-1}\vb{f}_l - \log|\vb{K}_m|\right]
        - \frac{NL}{2}\log(2\pi).
\label{eq:llhood}
\end{align}
We use a gradient-descent procedure to perform the optimization.

To predict the class label for a new matrix $\mathbf{F}_*$ or
equivalently, a new vector $\mathbf{f}_*$, and assuming equal prior
probabilities for each class, we compute the marginal likelihood
$p(\mathbf{F}_*|\mathbf{X}_*, \mathcal{C}_m, \bm{\theta}_m)$ for all $m$.
We predict as the correct class that one for which the
marginal likelihood is bigger.


\subsection{Discriminative Training}

In discriminative training, we search for the hyperparameters that
minimize some classification error measure for all classes simultaneously.
In this report, we chose to minimize the minimum classification error criterion as presented in
\cite{juangmce}. A soft version of the \{0,1\} loss function for classification can be written as
\begin{align}
    \ell_m(\vb{f}) &= \frac{1}{1 + \exp(-\gamma_1 d_m(\vb{f}) + \gamma_2)}, \label{eq:tmploss}
\end{align}
where $\gamma_1 > 0$ and $\gamma_2$ are user given
parameters, and $d_m(\vb{f})$ is the \emph{class misclassification measure}, given by
\begin{align}
   d_m(\vb{f}) &= -g_m(\vb{f}) + \log{\left[ \frac{1}{M-1}
    \underset{\underset{k\neq m}{\forall k}}
    {\sum} e^{g_k(\vb{f})\eta} \right]^{\frac{1}{\eta}}}, \label{eq:di}
\end{align}
where $\eta>0$, and $g_m(\vb{f}) = a\log{p(\vb{f}|\vb{X},
  \mathcal{C}_m, \bm{\theta}_m)} + b = a\log{\mathcal{N}(\vb{f}|\vb{0},\vb{K}_m)} + b$.
Parameters $a>0$, and $b$ are again defined by the user.  Expression $g_m(\vb{f})$ is an scaled and
translated version of the log marginal likelihood for the multi-output GP of class $m$. We scale the log
marginal likelihood to keep the value of $g_m(\vb{f})$ in a small
numerical range such that computing $\exp(g_k(\vb{f})\eta)$ does not
overflow the capacity of a double floating point number of a
computer. Parameters $\gamma_1$ and $\gamma_2$ in equation
\eqref{eq:tmploss} have the same role as $a$ and $b$ in $g_m(\vb{f})$, but the numerical problems are less severe
here and setting $\gamma_1=1$ and $\gamma_2=0$ usually works well.

Expression in equation \eqref{eq:di} converges to $-g_m(\vb{f}) +
\max_{\forall k: k \neq m}{g_k(\vb{f})}$ as $\eta$ tends to
infinity. For finite values of $\eta$, $d_m(\vb{f})$ is a
differentiable function. The value of $d_m(\vb{f})$ is negative if
$g_m(\vb{f})$ is greater than the ``maximum'' of $g_k(\vb{f})$, for
$k\ne m$. We expect this to be the case, if $\vb{f}$ truly belongs to
class $\mathcal{C}_m$. Therefore, expression $d_m(\vb{f})$ plays the
role of a discriminant function between $g_m(\vb{f})$ and the
``maximum'' of $g_k(\vb{f})$, with $k\neq m$.\footnote{We use quotes
  for the word maximum, since the true maximum is only achieved when
  $\eta$ tends to infinity.} The misclassification measure is a continuous function of
$\bm\Theta$, and attempts to model a decision rule. Notice that if $d_m(\vb{f})<0$ then
$\ell_m(\vb{f})$ goes to zero, and if $d_m(\vb{f})>0$ then
$\ell_m(\vb{f})$ goes to one, and that is the reason as why expression
\eqref{eq:tmploss} can be seen as a soft version of a \{0,1\} loss
function. The loss function
takes into account the class-conditional densities $p(\vb{f}|\vb{X},
\mathcal{C}_m, \bm{\theta}_m)$, for all classes, and thus, optimizing
$\ell_m(\mathbf{f})$ implies the optimization over the set
$\bm{\Theta}$.

Given some dataset $\{\mathbf{X}_m, \mathbf{F}_m\}_{m=1}^M$, the purpose is then to find the hyperparameters
$\bm{\Theta}$ that minimize the cost function that counts the number of misclassification errors in the dataset,
 \begin{equation}
     \mathcal{L}(\{\mathbf{X}_m\}_{m=1}^M, \{\mathbf{F}_m\}_{m=1}^M, \bm{\Theta}) =
\sum_{m=1}^{M}\sum_{l=1}^{L} \ell_m(\vb{f}_m^l).
\label{eq:mceloss}
 \end{equation}
We can compute the derivatives of equation \eqref{eq:mceloss} with respect to the hyperparameters $\bm{\Theta}$,
and then use a gradient optimization method to find the optimal hyperparameters for the minimum classification
error criterion. 




\subsubsection{Computational complexity}

Equation \eqref{eq:di} requires us to compute the sum over all possible classes to compute
the denominator. And to compute equation \eqref{eq:llhood}, we need to
invert the matrix $\vb{K}_m$ of dimension $DN \times DN$. The
computational complexity of each optimization step is then
$O(LMD^3N^3)$, this could be very slow for many applications.

In order to reduce computational complexity, in this report we resort
to low rank approximations for the covariance matrix appearing on the
likelihood model. In particular, we use the partially independent
training conditional (PITC) approximation, and the fully independent
training conditional (FITC) approximation, both approximations for
multi-output GPs \cite{alvarez2011computationally}.

These approximations reduce the complexity to $O(LMK^2DN)$, where $K$ is a
parameter specified by the user. The value of $K$ refers to the number of auxiliary input
variables used for performing the low rank approximations. The locations of these input variables
can be optimized withing the same optimization procedure used for finding the hyperparameters
$\bm{\Theta}$. For details, the reader is referred to
\cite{alvarez2011computationally}.

\section{Experimental Results}
\label{sec:results}
In the following sections, we show results for different experiments
that compare the following methods: hidden Markov models trained in a
generative way using the Baum-Welch algorithm \cite{RabinerHMM:1989},
hidden Markov models trained in a discriminative way using minimum
classification error \cite{juangmce}, multi-output GPs trained in a
generative way using maximum likelihood (this report), and multi-output
GPs trained in a discriminative way using minimum classification error
(this report). On section \ref{subs:exp1} we test the different
methods, for emotion classification from video sequences on the
Cohn-Kanade Database \cite{CKdatabase}. On section \ref{subs:exp2}, we compare the
methods for activity recognition (Running and walking) from video
sequences on the CMU MOCAP Database. On section \ref{subs:exp3}, we
use again the CMU MOCAP database to identify subjects from their
walking styles.  For this experiment, we also try different frame
rates for the training set and validation cameras to show how the
multi-output GP method adapts to this case. Finally on section
\ref{subs:exp4}, we show an example of face recognition from images. Our intention
here is to show our method on an example in which the dimensionality of the input
space, $p$, is greater than one.

For all the experiments, we assume that the HMMs have a Gaussian distribution per
state. The number of hidden states of a HMM are shown in each of the experiments
in parenthesis, for instance, HMM(q) means a HMM with $q$ hidden
states.

\subsection{Emotion Recognition from Sequence Images}
\label{subs:exp1}

\begin{figure}[ht]
\centering
\begin{subfigure}[t]{.31\textwidth}
  \includegraphics[width=\textwidth]{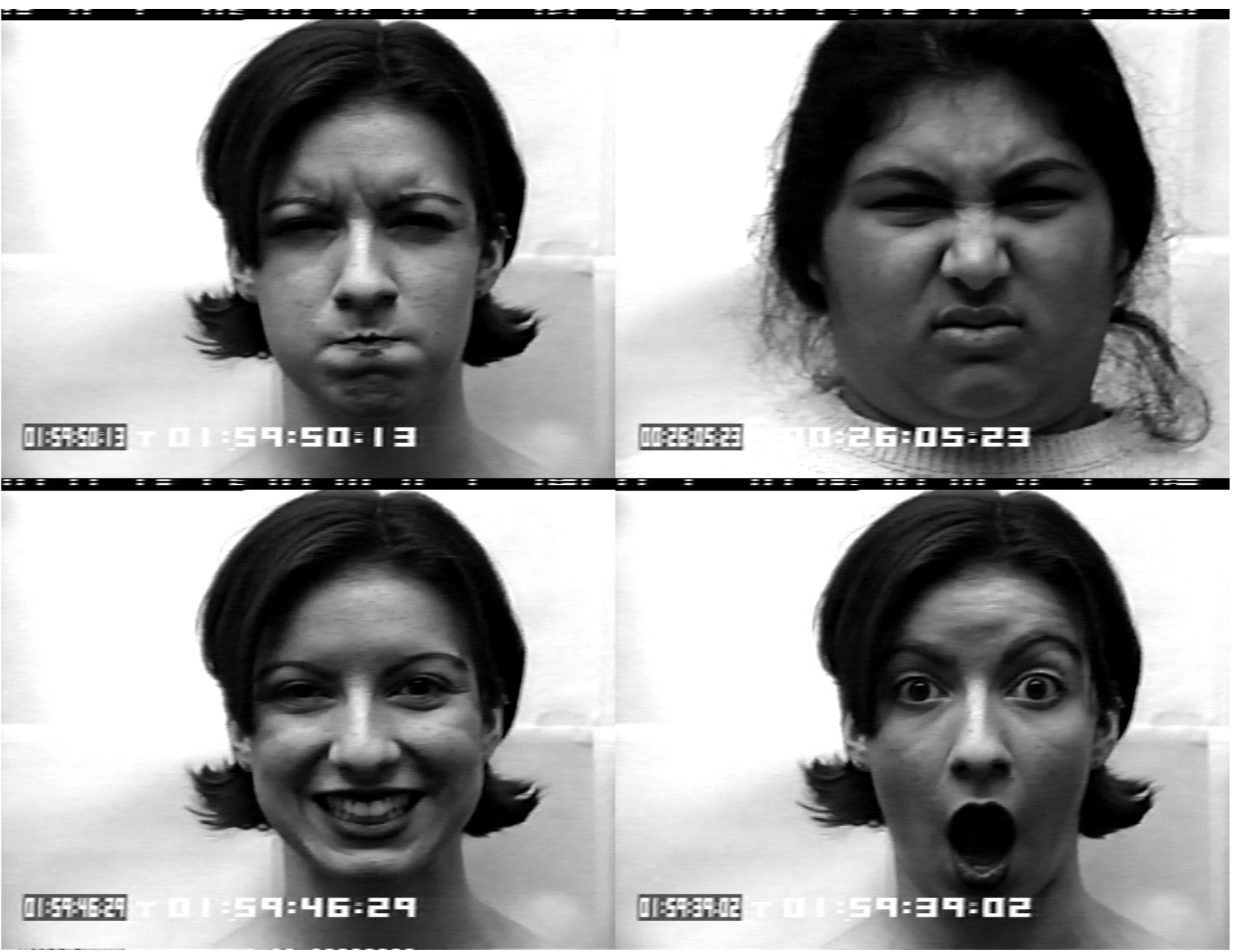}
  \caption{}
  \label{fig:faceCK}
\end{subfigure}
~
\begin{subfigure}[t]{.31\textwidth}
  \includegraphics[width=\textwidth]{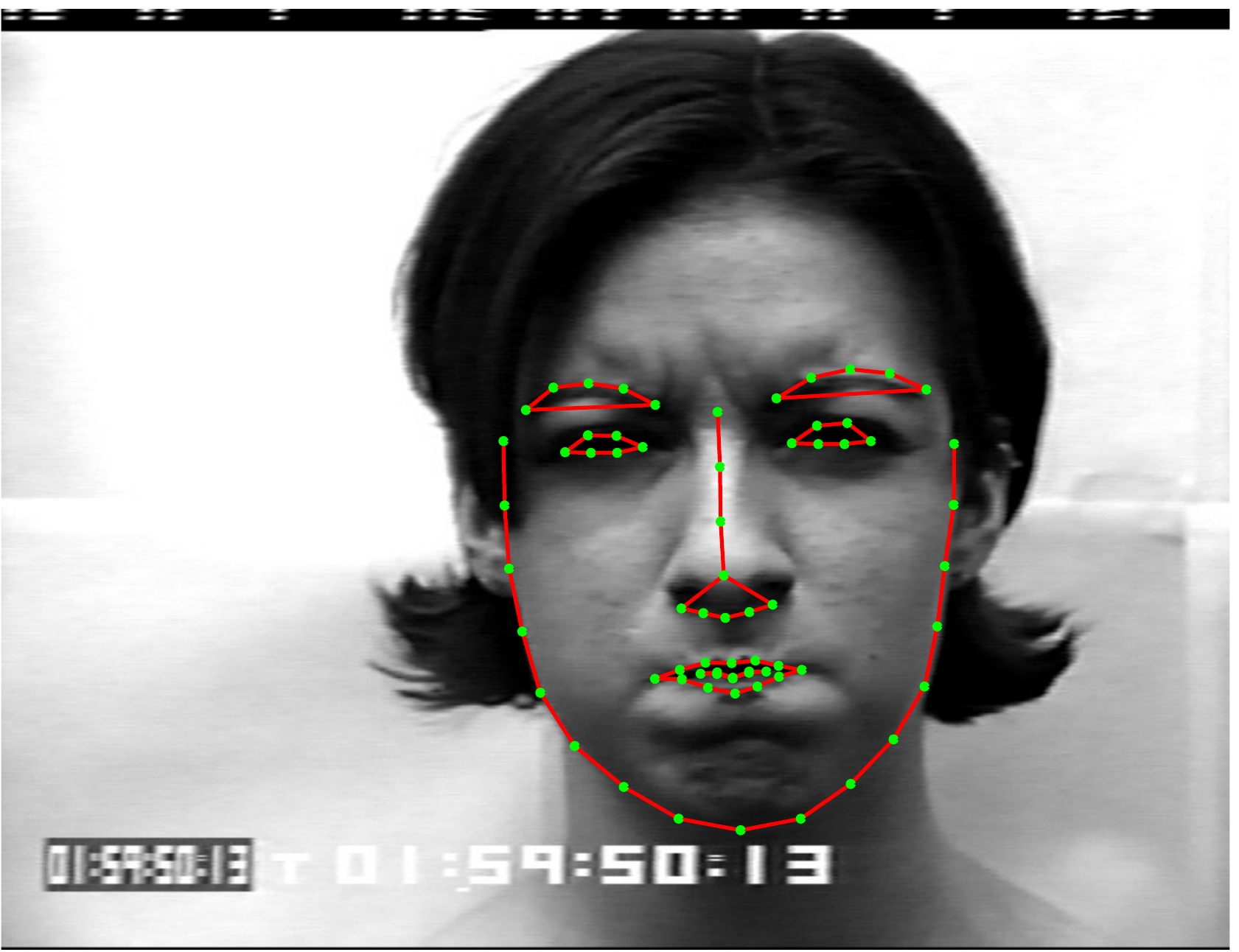}
  \caption{}
  \label{fig:faceShapeCK}
\end{subfigure}
~
\begin{subfigure}[t]{.31\textwidth}
  \centering
  \includegraphics[width=\textwidth]{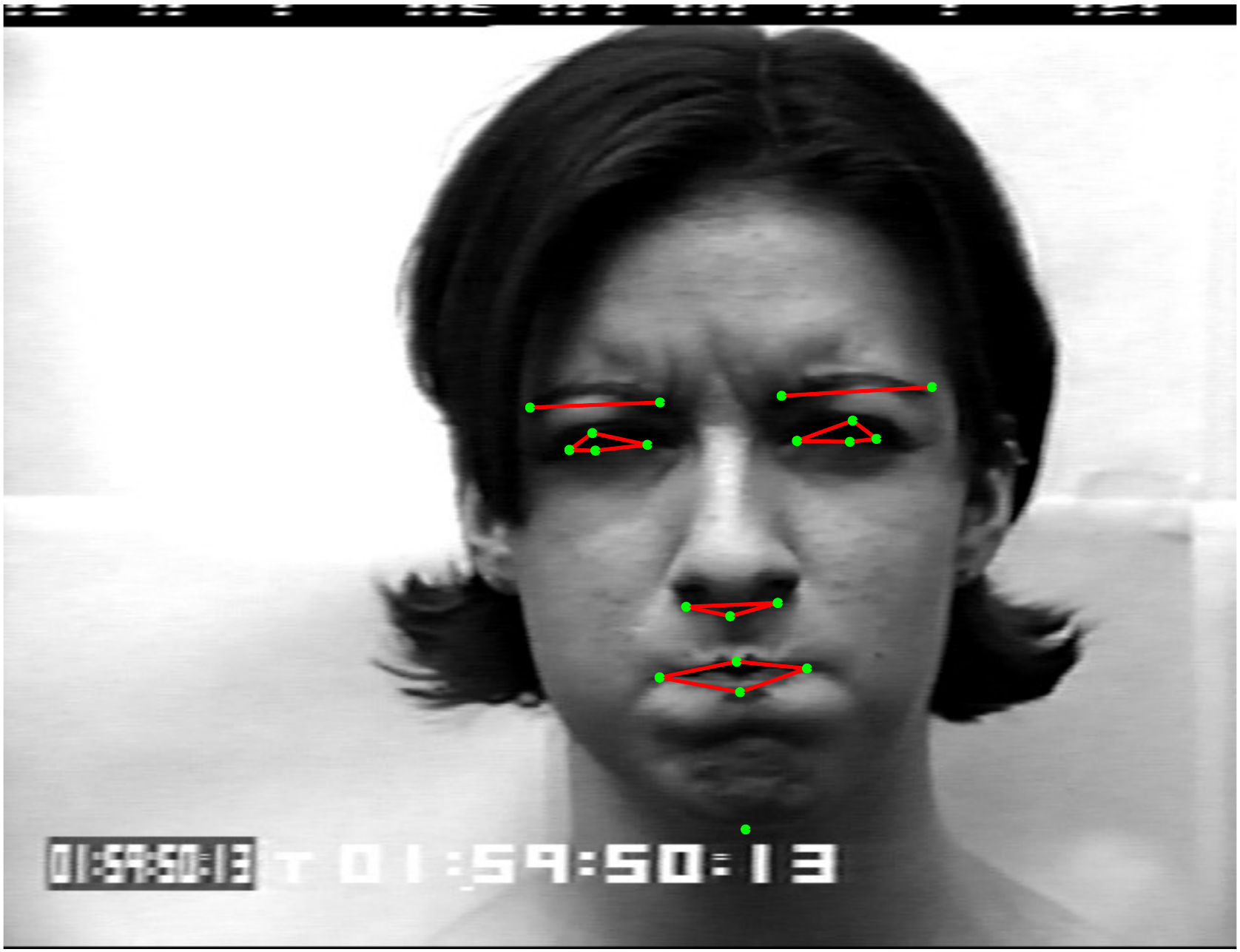}
  \caption{}
  \label{fig:faceShapeCKP}
\end{subfigure}

\caption{Cohn-Kanade emotion recognition example. \textbf{a)} Sample faces from CK database.
\textbf{b)} Facial landmarks provided by CK database \cite{CKdatabase}.
\textbf{c)} Key shape points according to \cite{Valstar12}}.
\label{fig:CKExp}
\end{figure}

For the first experiment,  we used the Cohn-Kanade Database
\cite{CKdatabase}. This database consists of processed videos of
people expressing emotions, starting from a neutral face and going to
the final emotion. The features are the positions $(x,y)$ of some
key-points or landmarks in the face of the person expressing the emotion. The
database consists on seven emotions. We used four emotions, those having
more than 40 realizations, namely, anger, disgust, happiness, and
surprise. This is $M=4$. Each instance consists on 68 landmarks evolving over time.
Figure \ref*{fig:CKExp} shows a description for the Cohn-Kanade facial
expression database. We employed 19 of
those 68 key points (see figure \ref{fig:faceShapeCK}), associated to the lips, and the eyes
among others, and that are thought to be more
relevant for emotion recognition, according to \cite{Valstar12}. Figure \ref{fig:faceShapeCKP}
shows these relevant features.

In this experiment, we model the coordinate $x$, and the coordinate
$y$ of each landmark, as a time series, this is, $x(t)$, and
$y(t)$. With $19$ landmarks, and two time series per landmark, we are
modeling multivariate time series of dimension $D= 38$. The length of
each time series $N$ was fixed in this first experiment to 71, using a
dynamic time warping algorithm for multiple signals \cite{Zhou12}. Our
matrices $\mathbf{F}_m^l\in\mathbb{R}^{38\times 71}$. For each class,
we have $L=40$ instances, and use 70\% for the training set, and 30\%
for validation set. We repeated the experiments five times. Each time,
we had different instances in the training set and the validation
set. We trained multi-output GPs with FITC and PITC approximations,
both fixing, and optimizing the auxiliary variables for the low rank
approximations. The number of auxiliary variables was $K=25$. When
not optimized, the auxiliary variables were uniformly placed along the
input space.

Accuracy results are shown in Table \ref{tab:dtw_results}. The table provides the mean, and
the standard deviation for the results of the five repetitions of the experiment. The star
symbol (*) on the method name means that the auxiliary input points
were optimized, otherwise the auxiliary points were fixed.

\begin{table}[t]
    \centering
    \begin{tabular}{cc@{\pmsep}cc@{\pmsep}c}
        \toprule
        Method & \multicolumn{2}{c}{Generative} & \multicolumn{2}{c}{MCE} \\
       \midrule
        FITC & 79.58 & 4.52 & 89.55 & 2.53 \\
        PITC & 69.16 & 16.82 & 87.07 & 5.77 \\
        FITC* & 79.16 & 3.29 & 89.16 & 6.49 \\
        PITC* & 70.83 & 16.21 & 85.82 & 3.73 \\
        HMM (3) & 85.80 & 7.26 & 84.15 & 9.60 \\
        HMM (5) & 79.00 & 3.74 & 87.91 & 4.27 \\
        HMM (7) & 70.80 & 8.87 & 91.66 & 6.08 \\
        \bottomrule
    \end{tabular}
    \caption{Classification accuracy (mean and standard deviation) for emotion
      recognition on the Cohn-Kanade database using dynamic time
      warping on the features. The asterisk in the table means that we also
      optimized the locations of the auxiliary variables for the low rank approximation.}
    \label{tab:dtw_results}
\end{table}

Gen and Disc refer to the generative and discriminative training, using either
the multi-output GP or the HMM. The table shows that for multi-output
GPs, discriminative training leads to better results than the
generative training. The table also shows results for HMM with 3, 5
and 7 hidden states respectively. Results for the HMM with generative training, and the
multi-output GP with generative training are within the same range, if we take into account
the standard deviation. Accuracies are also similar, when comparing the HMM trained with MCE,
and the multi-output GP trained with MCE. We experimentally show then that the multi-output GP
is as good as the HMM for emotion recognition.

\subsection{Activity Recognition With Motion Capture Data Set}
\label{subs:exp2}
\begin{figure}
    \centering
    \includegraphics[width=0.3\textwidth]{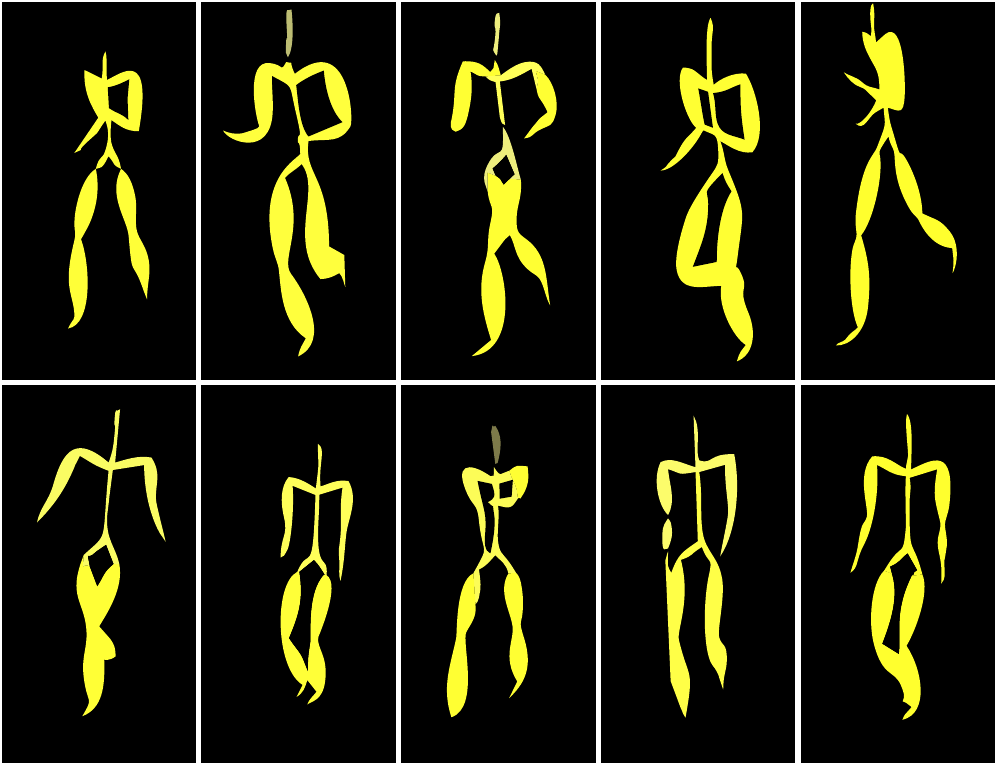}
    \captionof{figure}{Sample actions from MOCAP database, run action (above) and walk action (below). }
    \label{fig:ExpWalkRun}
\end{figure}

For the second experiment, we use a motion capture database (MOCAP)
to classify between walking and running actions. In MOCAP, the input
consists of the different angles between the bones of a 3D
skeleton. The camera used for the MOCAP database has a frame rate of
120 frames per second, but in this experiment we sub-sampled the
frames to $\frac{1}{8}$ of the original frame rate. Our motion capture
data set is from the CMU motion capture data base.\footnote{The CMU
  Graphics Lab Motion Capture Database was created with funding from
  NSF EIA-0196217 and is available at \url{http://mocap.
  cs.cmu.edu.}} We considered two different categories of movement:
running and walking. For running, we take subject 2 motion 3,
subject 9 motions 1$-$11, subject 16 motions 35, 36, 45, 46, 55, 56,
subject 35 motions 17$-$26, subject 127 motions 3, 6, 7, 8, subject
141 motions 1, 2, 3 34, subject 143 motions 1, 42, and for walking we
take subject 7 motions 1$-$11, subject 8 motions 1$-$10, subject 35
motions 1$-$11, subject 39 motions 1$-$10. Figure
\ref*{fig:ExpWalkRun} shows an example for activity recognition in
MOCAP database. In this example then, we have two classes, $M=2$, and $D=62$ time courses
of angles, modeled as a multi-variate time series. We also have $L_1 = 38$ for running, and
$L_2 = 42$ for walking.

Here again we compare the generative and discriminative approaches on
both our proposed model with FITC and PITC approximations and
HMMs. Again, we assume $K=25$. One important difference between the
experiment of section \ref{subs:exp1}, and this experiment is that we
are using the raw features here, whereas in the experiment of section
\ref{subs:exp1}, we first performed dynamic time warping to make that
all the time series have the same length. It means that for this experiment, $N_m^l$ actually
depends on the particular $m$, and $l$.

\begin{table}
    \centering
    \begin{tabular}{cc@{\pmsep}cc@{\pmsep}c}
        \toprule
        Method & \multicolumn{2}{c}{Generative} & \multicolumn{2}{c}{Discriminative}  \\
        \midrule
        FITC & 60.68 & 3.98 & 95.71 & 2.98 \\
        PITC & 76.40 & 12.38 & 93.56 & 5.29 \\
        FITC* & 58.90 & 0.00 & 96.78 & 1.95 \\
        PITC* & 69.28 & 15.28 & 84.90 & 11.33 \\
        HMM (3) & 96.70 & 2.77 & 97.95 & 2.23 \\
        HMM (5) & 94.69 & 4.36 & 96.32 & 0.82 \\
        HMM (7) & 92.24 & 4.49 & 99.77 & 0.99 \\
        \bottomrule
    \end{tabular}
    \captionof{table}{Classification accuracy rates (mean and standard deviation) for activity identification on the
    CMU-MOCAP database.}
    \label{tab:mocap1}
\end{table}

The results are shown in Table \ref{tab:mocap1} for five repeats of
the experiment.  For each repeat, we used $15$ instances for training, and the rest in each class, for
validation. Again the results are comparable with the results of
the HMM within the standard deviation. As before, the discriminative
approach shows in general better results than the generative approach.

Experiments in this section and section \ref{subs:exp1}, show that
multi-output GPs exhibit similar performances to HMM, when used for
pattern recognition of different types of multi-variate time series.

\subsection{Subject Identification on a Motion Capture Data Set}
\label{subs:exp3}

For the third experiment we took again the CMU MOCAP database but instead of classifying between
different actions, we recognized subjects by their walking styles. We considered three different
subjects exhibiting walk movements. To perform the identification we took subject 7 motions
1,2,3,6,7,8,9,10, subject 8 motions 1,2,3,5,6,8,9,10, and subject 35 motions 1$-$8. Then for each
subject we took four instances for training and other four repetitions for validation.
Figure 3 shows an example for subject identification in the CMU-MOCAP database.
We then have $M=3$, $D=62$, $L=8$, and the length for each instance, $N_m^l$, was variable.

For this experiment, we supposed the scenario where the frame rate for the motions used in training
could be different from the frame rate for the motions used in testing. This configuration simulates
the scenario where cameras with different recording rates are used to keep track of human activities.
Notice that HMMs are not supposed to adapt well to this scenario, since the Markov assumption is that the
current state depends only on the previous state. However, the Gaussian process captures
the dependencies of any order, and encodes those dependencies in the kernel function, which is a continuous
function of the input variables. Thus, we can evaluate the GP for any set of input points, at the testing
stage, without the need to train the whole model again.

\begin{figure}
  \begin{center}
  \includegraphics[width=0.26\textwidth]{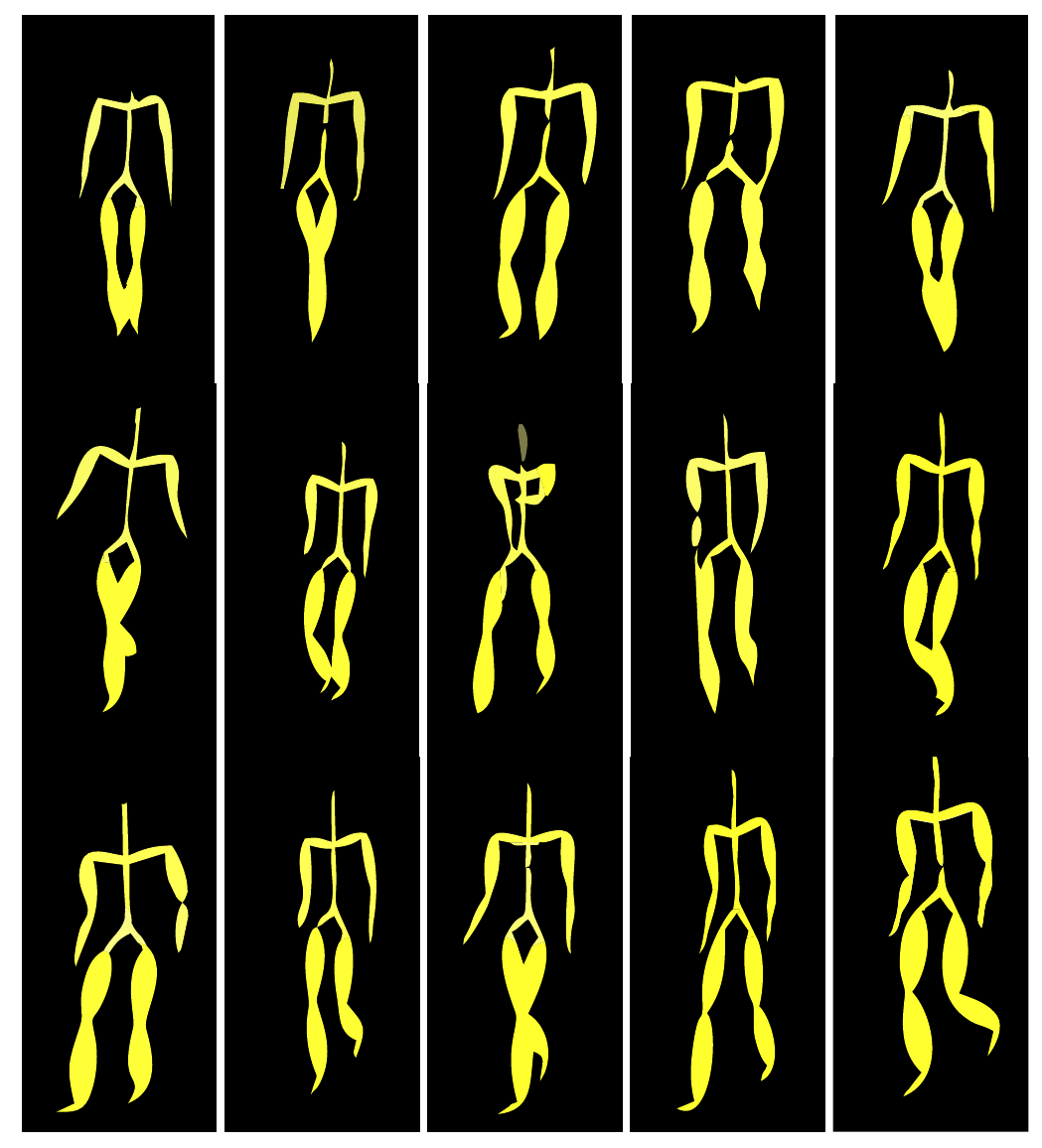}
  \caption{Sample walk actions from CMU MOCAP database, subjects 7, 8 and 35.}
  \end{center}
  \label{fig:ExpWalkAction}
\end{figure}

Table \ref{tab:decimation} shows the results of this experiment. In
the table, we study three different scenarios: one for which the frame
rate in the training instances was slower than in the validation
instances, one for which it was faster, and one for which it was the
same. We manipulate the frame rates by decimating in training (DT),
and decimating in validation (DV). For example, a decimation of
$\frac{1}{16}$ means that one of each 16 frames of the original time
series is taken. When the validation frame rate is faster than the
training frame rate (column Faster in Table \ref{tab:decimation}), the
performance of the multi-output GP is clearly superior to the one
exhibited by the HMM, both for the generative and the discriminative
approaches. When the validation frame rate is slower or equal than the
training frame rate (columns Slower and Equal), we could say that the
performances are similar (within the standard deviation) for
multi-output GPs and HMM, if they are trained with MCE. If the models are trained generatively,
the multi-output GP outperforms the HMM. Although the results for the HMM in Table \ref{tab:decimation}
were obtained fixing the number of states to seven, we also performed experiments for three and five states,
obtaining similar results. This experiment shows an example, where our model is clearly useful to solve
a problem that a HMM does not solve satisfactorily.

\begin{table}[H]
    \centering
        \begin{tabular}{c|c@{\pmsep}c|c@{\pmsep}c|c@{\pmsep}c}
        \toprule
        Method & \multicolumn{2}{c}{Faster} &
        \multicolumn{2}{c}{Slower} &
        \multicolumn{2}{c}{Equal} \\
            \midrule
            FITC Gen & 93.28 & 3.76 & 94.96 & 4.60 & 94.96 & 4.60 \\
            PITC Gen & 93.28 & 3.76 & 94.96 & 4.60 & 94.96 & 4.60 \\
            FITC MCE & 94.96 & 4.60 & 89.96 & 9.12 & 89.96 & 9.12 \\
            PITC MCE & 94.96 & 4.60 & 93.28 & 3.76 & 88.32  & 12.63 \\
            HMM Gen (7) & 33.33 & 0.00 & 36.40 & 12.56 & 81.60 & 6.98 \\
            HMM MCE (7) & 83.33 & 16.6 & 94.90 & 4.60 & 100.00 & 0.00 \\
            \bottomrule
        \end{tabular}
        \caption{Classification accuracy rates (mean and standard deviation) of
      subject identification by his walking style on the CMU-MOCAP database. The
      first column (Faster) shows the results for when the validation
      camera is 4 times faster than the training camera (DT =
      $\frac{1}{32}$, DV = $\frac{1}{8}$). The second column (Slower)
      shows the results for when the validation camera is 4 times
      slower than the training camera (DT = $\frac{1}{8}$,
      DV=$\frac{1}{32}$).  The last column (Equal) shows the results
      for when both the validation and training cameras have the same
      frequency (DT = $\frac{1}{8}$, DV=$\frac{1}{8}$).  }
    \label{tab:decimation}
\end{table}

\subsection{Face Recognition}
\label{subs:exp4}

\begin{figure}[H]
    \centering
    \includegraphics[width=0.29\textwidth]{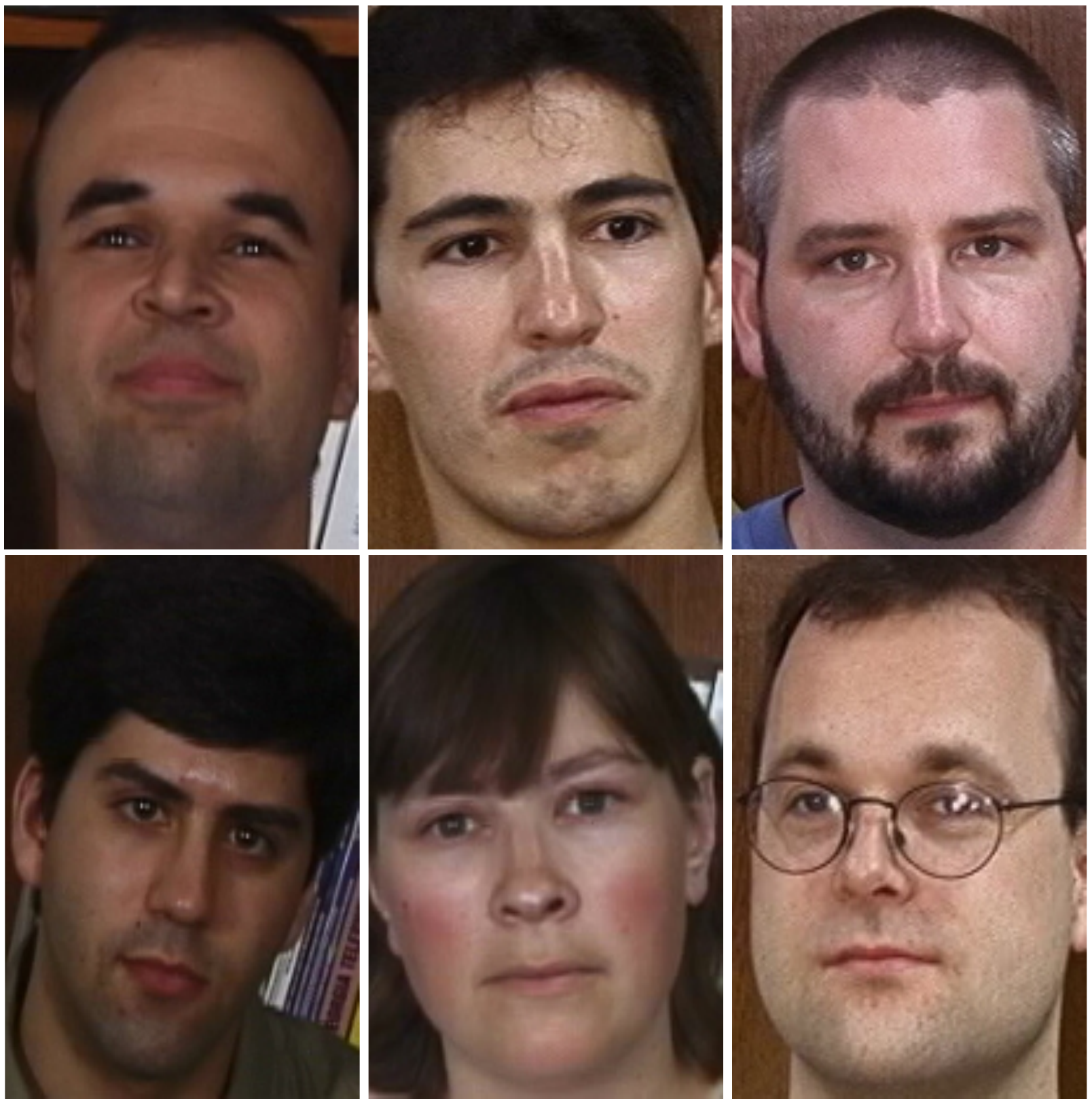}
    \captionof{figure}{Sample faces from Georgia Tech Face database.}
    \label{fig:faceRec}
\end{figure}

The goal of the fourth experiment is to show an example where the
vector-valued function is dependent on input variables with dimensionality greater than one,
functions of multi-dimensional
inputs $(f(\vb{x}_0),f(\vb{x}_1),...,f(\vb{x}_n))$ like space. The
HMMs as used here are not easily generalized in this case and, thus, we do not present
results with HMMs for this experiment. In this problem we work with
face recognition from pictures of the Georgia Tech
database \footnote{Georgia Tech Face Database,
  \url{http://www.anefian.com/research/face_reco.htm}}. This database,
contains images of 50 subjects stored in JPEG format with $640\times 480$
pixel resolution. For each individual 15 color images were taken,
considering variations in illumination, facial expression, face
orientations and appearance (presence of faces using glasses). Figure
\ref{fig:faceRec} shows an example for the Georgia Tech Face database.

Here we did two experiments. The first experiment was carried out
taking 5 subjects of the Georgia Tech database that did not have
glasses. For the second experiment we took another 5 subjects of the
database that had glasses.  In both experiments, each image was
divided in a given number of regions of equal aspect ratio. For each
region $n$ we computed its centroid $\vb{x}_n$ and a texture vector
$\vb{f}_n$. Notice that this can be directly modeled by a multi-output
GP where the input vectors $\vb{x}_n$ are two dimensional.

\begin{table}
    \centering
    \begin{tabular}{c c@{\pmsep}c c@{\pmsep}c}
        \toprule
        Method & \multicolumn{2}{c}{Gen} & \multicolumn{2}{c}{Disc} \\
        \midrule
        FITC & 61.57 & 3.50 & 86.84 & 0.01 \\
        PITC & 64.72 & 2.34 & 95.78 & 8.03 \\
        FITC*& 66.71 & 3.82 & 96.84 & 7.06 \\
        PITC*& 73.68 & 5.88 & 96.30 & 3.00 \\
        \bottomrule
    \end{tabular}
    \caption{Recognition accuracy (mean and standard deviation) for faces without glasses using a grid of size BX=4, BY=7.}
    \label{tab:faces_noglass0}
\end{table}

\begin{table}
    \centering
    \begin{tabular}{c c@{\pmsep}c c@{\pmsep}c}
        \toprule
        Method & \multicolumn{2}{c}{Gen} & \multicolumn{2}{c}{Disc} \\
        \midrule
        FITC & 51.57 & 3.5 & 88.42 & 2.35 \\
        PITC & 69.47 & 3.53 & 83.68 & 4.30 \\
        FITC*& 56.80 & 2.44 & 86.84 & 0.01 \\
        PITC*& 62.10 & 8.24 & 87.36 & 1.17 \\
        \bottomrule
    \end{tabular}
    \caption{Recognition accuracy (mean and standard deviation) for faces without glasses using a grid of size BX=6, BY=7.}
    \label{tab:faces_noglass1}
\end{table}

\begin{table}
    \centering
    \begin{tabular}{c c@{\pmsep}c c@{\pmsep}c}
        \toprule
        Method & \multicolumn{2}{c}{Gen} & \multicolumn{2}{c}{Disc} \\
        \midrule
        FITC & 54.73 & 6.55 & 81.57 & 3.7 \\
        PITC & 64.21 & 9.41 & 81.57 & 7.2 \\
        FITC* & 60.53 & 0.02 & 90.52 & 9.41 \\
        PITC* & 69.47 & 9.41 & 77.36 & 8.24 \\
        \bottomrule
    \end{tabular}
    \caption{Recognition accuracy (mean and standard deviation) for faces with glasses using a grid of BX=4, BY=7.}
    \label{tab:faces_glass1}
\end{table}

\begin{table}
    \centering
    \begin{tabular}{c c@{\pmsep}c c@{\pmsep}c}
        \toprule
        Method & \multicolumn{2}{c}{Gen} & \multicolumn{2}{c}{Disc}  \\
        \midrule
        FITC & 42.1 & 0.02 & 93.68 & 2.35 \\
        PITC & 35.78 & 2.35 & 86.84 & 0.01 \\
        FITC* & 72.6 & 5.45 & 86.84 & 0.01 \\
        PITC* & 48.42 & 2.35 & 89.47 & 0.01 \\
        \bottomrule
    \end{tabular}
    \caption{Recognition accuracy (mean and standard deviation) for faces with glasses using a grid of BX=6, BY=7.}
    \label{tab:faces_glass2}
\end{table}

Tables \ref{tab:faces_noglass0}, \ref{tab:faces_noglass1},
\ref{tab:faces_glass1} and \ref{tab:faces_glass2} show the results of
this experiment with the discriminative and the generative training
approaches. The number of divisions in the X and Y coordinates are BX
and BY respectively. The features extracted from each block are mean
RGB values and Segmentation-based Fractal Texture Analysis (SFTA)
\cite{SFTA} of each block. The SFTA algorithm extracts a feature
vector from each region by decomposing it into a set of binary images,
and then computing a scalar measure based on fractal symmetry for each
of those binary images.

The results show high accuracy in the recognition process in both
schemes (Faces with glasses and faces without glasses) when using
discriminative training. For all the settings, the results of the discriminative
training method are better than the results of the generative training
method. This experiment shows the versatility of the multi-output
Gaussian process to work in applications that go beyond time series
classification.

\section{Conclusions}
\label{sec:conclusions}

In this report, we advocated the use of multi-output GPs as generative models
for vector-valued random fields. We showed how to estimate the hyperparameters
of the multi-output GP in a generative way and in a discriminative way, and
through different experiments we demonstrated that the performance of our
framework is equal or better than its natural competitor, a HMM.

For future work, we would like to study the performance of the framework using
alternative discriminative criteria, like Maximum Mutual Information (MMI) using
gradient optimization or
Conditional Expectation Maximization \cite{Jebara:book:2004}. We would also
like to try practical applications for which there is the need to classify
vector-valued functions with higher dimensionality input spaces. Computational
complexity is still an issue, we would like to implement alternative efficient
methods for training the multi-output GPs \cite{Hensman:GP:BigData:2013}.

\section*{Acknowledgments}

SGG acknowledges the support from ``Beca Jorge Roa Mart\'inez'' from
Universidad Tecnol\'ogica de Pereira, Colombia.

\bibliographystyle{plainnat}
\bibliography{discriminative}

\end{document}